\documentclass{article}

\usepackage{arxiv}

\usepackage[utf8]{inputenc} 
\usepackage[T1]{fontenc}    
\usepackage{hyperref}       
\usepackage{url}            
\usepackage{booktabs}       
\usepackage{amsfonts,amsmath,amssymb}       
\usepackage{nicefrac}       
\usepackage{microtype}      
\usepackage{bm}
\usepackage{graphicx}
\usepackage{algorithmic}
\usepackage[ruled,vlined]{algorithm2e}
\usepackage{caption} 
\title{Learn to Design the Heuristics for Vehicle Routing Problem}

\author{
  Lei Gao \And Mingxiang Chen \And Qichang Chen \And Ganzhong Luo \And Nuoyi Zhu \And Zhixin Liu\\
  WaterMirror Inc.\\
  Shenzhen, China \\
  \texttt{liuzhixin@watermirror.ai} \\
}

\begin{document}
\maketitle

\begin{abstract}
This paper presents an approach to learn the local-search heuristics that iteratively improves the solution of Vehicle Routing Problem (VRP). A local-search heuristics is composed of a \textit{destroy} operator that destructs a candidate solution, and a following \textit{repair} operator that rebuilds the destructed one into a new one. The proposed neural network, as trained through actor-critic framework, consists of an encoder in form of a modified version of Graph Attention Network where node embeddings and edge embeddings are integrated, and a GRU-based decoder rendering a pair of destroy and repair operators. Experiment results show that it outperforms both the traditional heuristics algorithms and the existing neural combinatorial optimization for VRP on medium-scale data set, and is able to tackle the large-scale data set (e.g., over 400 nodes) which is a considerable challenge in this area. Moreover, the need for expertise and handcrafted heuristics design is eliminated due to the fact that the proposed network learns to design the heuristics with a better performance. Our implementation is available online.\footnote{https://github.com/water-mirror/NeuLNS}
\end{abstract}


\section{Introduction}
Combinatorial optimization is an active research area in applied mathematics and Operations Research, with a wide range of applications such as supply chain management, aviation planning, urban transportation, power industry, etc. It answers the question of "what is the optimal solution of an objective function with certain constraints". Vehicle Routing Problem (VRP) is one of the most important and challenging topics in combinatorial optimization. It aims at finding the optimal set of routes for a fleet of vehicles to deliver goods from depots to customers (nodes). The objective function of VRP is to minimize the total routing cost in terms of total mileage and/or size of fleet. In practice, the basic VRP is extended with constraints, for instance, on the time window, on the vehicle capacities, or on the time window of pick-up and delivery pairs. Finding the optimal solution of VRP is NP-hard, and exhaustive search is not tractable even with moderate number of nodes (e.g. 100). Sophisticated \textit{exact} algorithms like Branch-and-Bound and Column Generation are introduced to find the optimal solution, but still with heavy computational load that poses great challenge for large scale problems and when response time is crucial. Alternatively commercial solvers tent to use heuristic approaches to find a close-to-optimal solution for medium-to-large scale problems within a relatively short computational period. Very Large-scale Neighborhood Search (VLNS) is the major heuristic framework by iteratively transforming current solution into another feasible one in the \textit{neighborhood} for purpose of exploration and hence escaping from a local minima, where \textit{neighborhood} is defined as a set of solutions \textit{similar} to the current solution only with limited modifications. Intensive research on handcrafted heuristics design has been conducted in recent decades \cite{Hemmelmayr2012AnAL,Kytjoki2007AnEV,Cordeau2005TabuSH,Du2012CombiningNN,Gror2010ALO}.

Heuristics design employs notably the intuition from experts, making it a promising area to be integrated, and hence be automated and augmented, by machine learning. \cite{DBLP:journals/corr/abs-1811-06128} provides a survey of the generic methodology to integrate machine learning and combinatorial optimization. 
\cite{Bello2016NeuralCO} uses a Seq2Seq framework \cite{Sutskever2014SequenceTS} to predict a distribution over permutations of nodes to be visited for Traveling Salesman Problem (TSP), trained by Policy Gradient method\cite{Sutton1999PolicyGM}. \cite{Nazari2018ReinforcementLF} uses graph embedding together with Attention mechanism as the encoder, and RNN as the decoder to generate a sequence of consecutive actions for the VRP. \cite{Kool2018AttentionLT,Kool2019Buy4R} proposed an multi-head attention model to solve routing problems such as TSP and VRP. These three approaches share some common methodologies, despite their different model designs, as: 1) encoder-decoder architecture with attention is applied to generate sequential decisions that construct the routes, 2) embedding is essential on the encoder side to incorporate information about the nodes and the demands, and 3) network is trained by Policy Gradient. \cite{Ma2019CombinatorialOB} applies Pointer Network \cite{Vinyals2015PointerN} and hierarchical reinforcement learning \cite{Kulkarni2016HierarchicalDR} to solve the VRP with constraints, following the similar methodology above, in addition with the idea of hierarchical reinforcement learning to decouple the constraints from the objective function. As depicted in (13) of this paper the reward is the total cost plus the penalty, and this reward yields to Lagrange relaxation which is only a lower bound of the original VRP. Instead of generating the routes directly as the sequential output of the decoder, \cite{Chen2018LearningTP} proposed a reinforcement learning method to improve the existing solution by iteratively applying the single removal-then-insertion (i.e. 1-exchange) heuristic in the framework of VLNS. This approach integrates deep learning with heuristic algorithm to improve the effectiveness and quality of the solution, but the neural network only helps the 1-exchange policy as one concrete heuristic.

In this paper, we propose a novel approach to learn how to design the universal heuristics rather than a concrete one (for instance, 1-exchange in \cite{Chen2018LearningTP}), that iteratively improves an initial solution of VRP until it converges to a feasible solution close to optimal. Our approach is inspired by the research work we briefed above, for instance, the idea of embedding and encoder-decoder architecture trained by actor-critic, but with novel design. The contributions of the proposed approach are three-fold:
\begin{itemize}
	\item The proposed approach learns to design the generic heuristics without supervised training set from experts, and the learned heuristics outperforms traditional handcrafted ones and other neural combinatorial optimization algorithms. To best of our knowledge, our approach is the first one to design generic heuristics by machine learning, implying the need for handcrafted heuristics design is eliminated.
	\item In our approach the embedding consists of node embedding and edge embedding, both of which are integrated by a modified version of Graph Attention Network (GAT) \cite{velivckovic2017graph} as a joint representation of the demands and the topology. Therefore it is compatible for non-Euclidean space, where traveling distance does not equal to the Euclidean distance between any two nodes.
	\item The proposed approach is able to solve VRP with large scale, for example, more than 400 nodes, in a relatively short period. As far as we know that other neural combinatorial optimization algorithms are not able to get a close-to-optimal solution within acceptable period for the VRP with this level of scale. 
\end{itemize}

This paper is organized as follows. First we discuss the background of VRP and VLNS in Section 2, then introduce our method along with the new design of neural network in Section 3. The experiments setting, training details, and experiments results are presented in Section 4. The final conclusion and future work are discussed in Section 5.

\section{Background}

\subsection{VRP}
VRP is defined on an directed graph $\mathcal{G}=(\mathcal{N},\mathcal{A})$ where $i\in\mathcal{N}=\{0,1,...,N\}$ represents the $i$-th node for customer if $i>0$ or depot if $i=0$, and $a_{i,j}\in\mathcal{A},i,j\in\mathcal{N},i\neq j$ represents the arc from node $i$ to $j$. The demand of goods to be delivered to node $i$ is $q_i$, the service time window on node $i$ starts from $s_i$ and ends at $e_i$. The goal of VRP is to find one or several Hamiltonian cycles (a cyclic tour where each vertex is visited exactly once) that 1) fulfill the demands $q_i, i\in\mathcal{N}$, 2) served by part or all of $K$ vehicles, and 3) with constraints about the service time windows, the capacities of vehicles, etc. The set of Hamiltonian cycles will be referred as a \textit{solution} hereafter.

\subsection{Very Large-scale Neighborhood Search (VLNS)}
\textit{Neighborhood Search} is a framework to iteratively search the neighborhood of current solution and then apply the local minima as a substitution. In this extent the solution will be converging to a close-to-optimal one gradually. The neighborhood of solution $x$ is defined as $N(x)=\{x^\prime=\mathcal{H}(x^{\prime\prime}), x^{\prime\prime}\in x\bigcup N(x)\}$, where $\mathcal{H}(x)$ denotes the heuristics operator that explores the alternative solutions similar to $x$. A good design of heuristics operator $\mathcal{H}$ explicitly leads to better exploration effectiveness and converging efficiency. The heuristics can be as simple as exchanging $k$ nodes across the routes, such as 1-exchange where one node is picked-then-rewritten, or 2-exchange that swaps two nodes. The heuristics defined in this simple manner leads to a small scope of neighborhood and a limited exploration potential. With more sophisticated VLNS framework \cite{LNS_Shaw} the neighborhood is defined implicitly by a pair of \textit{destroy} and \textit{repair} operators. The destroy operator destructs part of the current solution by removing several nodes out of the routes, and the repair operator rebuilds the destructed solution by sequentially inserting the removed nodes back. The destroy operator typically removes a large part of the solution, say, up to 10\% or more, at different positions with stochasticity in each cycle, extending the neighborhood to a much larger scope. The repair operator inserts the removed nodes following a least-cost principle, and the solution quality varies with insertion order. Our approach targets at learning the destroy operator (in terms of the removal pattern) and repair operator (in terms of the insertion order) through a GAT-based encoder and a GRU-based decoder trained by actor-critic framework.

\section{Proposed Model}
\subsection{Embeddings}
The node information and topology are included in the primitive embedding for nodes and for edges, respectively. The primitive node embedding for $i$-th node is a 8-dimensional vector $n_i$ with each element as:
1) service starting time $s_i$, 2) service ending time $e_i$, 3) demand $q_i$, 4) total demands of the corresponding route, 5) sum up of demands of the corresponding route till this node, 6) total traveling distance along this route till this node, 7) traveling time along this route till this node, and 8) the possible forward shift defined in \cite{Savelsbergh1990API}.
Meanwhile, the primitive edge embedding for the arc $a_{i,j}$ connecting node $n_i$ and $n_j$ is a 2-dimensional vector $e_{i,j}$ consisting of the traveling distance along $a_{i,j}$, and a binary indicator about whether this arc is part of solution or not.

The elements of both embeddings are normalized. The normalized primitive embeddings are further integrated by a modified version of Graph Attention Network (\textit{EGATE}, or Element-wise GAT with Edge-embedding) that will be introduced in the next section.

\subsection{The Encoder}

The Graph Attention Network (GAT) \cite{velivckovic2017graph} is the neural network architecture with great representation power for graph topology by propagating node information via attention mechanism. For VRP in the form of directed graph $\mathcal{G}=(\mathcal{N},\mathcal{A})$, the arc $a_{i,j}\in\mathcal{A}$ carries information such as the cost and driving distance per trip, whereas GAT only integrates information carried by nodes $i\in\mathcal{N}$. This fact motivates us to introduce the EGATE as a modified version of GAT, where the node embeddings are not only updated with other nodes via attention weights, as proposed by the original GAT, but also are updated with the arc information represented by edge embedding $e_{i,j}$. Therefore both node set $\mathcal{N}$ and arc set $\mathcal{A}$ contribute to the attention weights for each node, as described mathematically in the following.

The primitive embeddings for node and edge introduced in previous section are firstly extended by a fully connection layer
\begin{eqnarray}
\tilde{n}_i&=&W_{n}\ast n_i\\
\tilde{e}_{i,j}&=&W_{\text{edge}}\ast e_{i,j}
\end{eqnarray}
The inputs to EGATE are the extended embeddings ($\tilde{n}_i,\tilde{e}_{i,j}$), and the output is an updated node embedding $n_{\text{EGATE},i}$. 
At first the attention weight vector for pair $i,j$ is calculated as:
\begin{eqnarray}
h_{\text{concat},ij}&=&\text{concat}(\tilde{n}_i,\tilde{n}_j,\tilde{e}_{i,j})\\
w_{i,j}&=&\text{LeakyReLU}(W_L\ast h_{\text{concat},ij})\\
\tilde{w}_{i,j}&=&\frac{\exp(w_{i,j})}{\sum_j \exp(w_{i,j})}
\end{eqnarray}

\begin{figure}[hbt!]
    \begin{center}
        \begin{minipage}{0.42\linewidth}
            \centering
            \includegraphics[width=0.9\columnwidth]{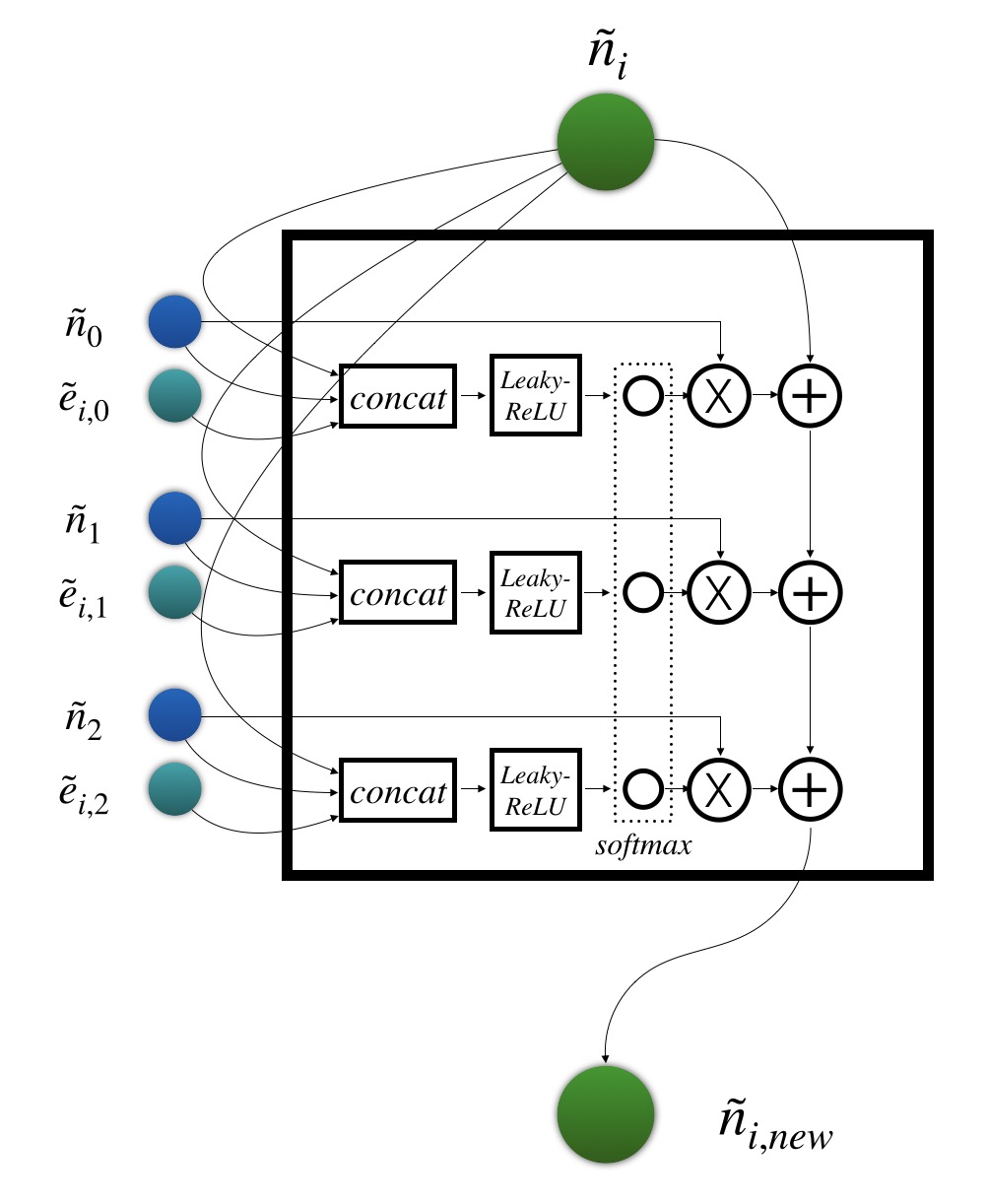}
        \end{minipage}\hfill
        \begin{minipage}{0.56\linewidth}
            \centering
            \includegraphics[width=\columnwidth]{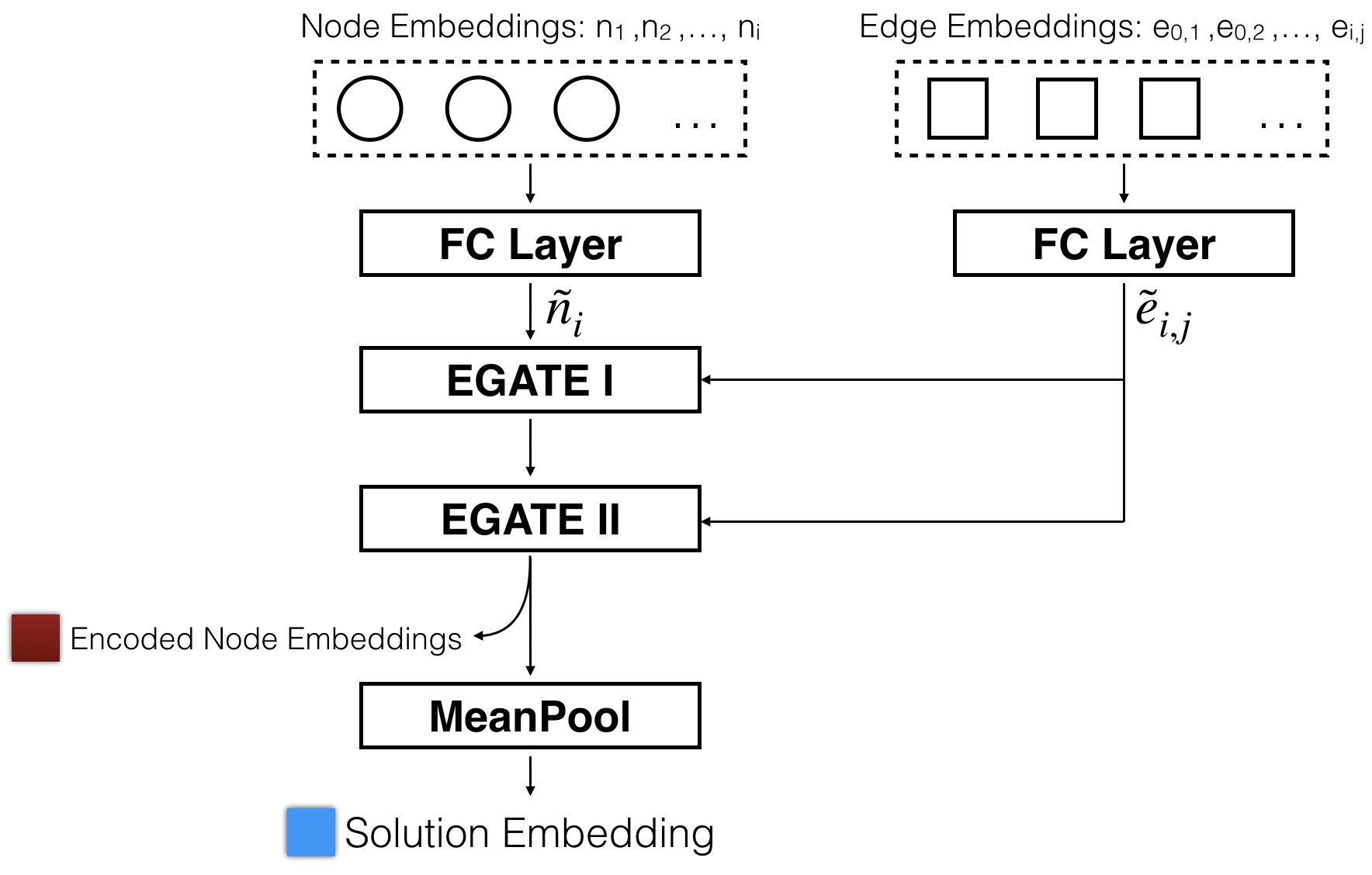}
        \end{minipage}
    \end{center}
    \caption{Left: The structure of Element-wise GAT with Edge Embedding (EGATE). Right: The network structure of the encoder.}\label{fig:egate_encoder}
\end{figure}

Then each node's embedding is updated to produce the output of EGATE via attention mechanism similar to GAT: 
\begin{equation}
n_{\text{EGATE},i}=\tilde{n}_i+\sum_j \tilde{w}_{i,j}\otimes \tilde{n}_j\label{eqn_egate_embedding}
\end{equation}
where $\otimes$ represents element-wise multiplication between vectors, and $W_n, W_{\text{edge}}, W_L$ are weight matrices to be learned. The output of EGATE $n_{\text{EGATE},i}$ represents the node's feature by propagating the information from other nodes as well as from arcs. Therefore EGATE provides a richer expression to the graph topology. It is worth noting that following the \textit{masked-attention} principle of GAT, EGATE allows some edge embeddings to be excluded in the information propagation by selectively masking them. For example, the arcs that violate the constraints on traveling time are excluded to reduce the computational load. 
The structure of EGATE is depicted in fig. \ref{fig:egate_encoder}.

Given the definition of single-layer EGATE, in practice we can cascade multiple EGATE layers so that each node has a wider \textit{receptive field}, hence a greater chance to exchange information with other nodes across longer distance. The input $\tilde{n}_i$ to higher EGATE layer is the output $n_{\text{EGATE},i}$ of lower EGATE layer, while the extended edge embedding $\tilde{e}_{i,j}$ keeps unchanged for all EGATE layers. Different masking strategies for either nodes or arcs can be applied to different EGATE layers.

The outputs of the highest EGATE layer are the final version of node embeddings. They are further fed into a mean-pooling layer to produce the final output of the encoder that representing the entire solution. 

The network structure of the encoder is depicted in fig. \ref{fig:egate_encoder}, with two cascaded EGATE layers as an example.

\subsection{The Decoder}
We design the decoder to learn the VLNS heuristics in a way such that the destroy operator produces a subset of nodes as removal candidates, which are inserted back by the repair operator in a strict order, and each insertion follows the least-cost principle yielding the minimum cost at current stage. In another word, the destroy operator we learned is to define a set of nodes, and the repair operator we learned is to reshape this set into an ordered list. Therefore the heuristics operator $\mathcal{H}$ at each iteration produces an ordered list directly as:
\begin{equation}
\mathcal{H}=\pi([\eta_1,\eta_2,...,\eta_M])\label{eqn_heuristic_form}
\end{equation}
where $\eta_m\in\mathcal{N}, m=1,2,...,M$ are the candidate nodes to be removed, $[\cdot]$ denotes a list with an order, and $\pi$ is the joint probability. (\ref{eqn_heuristic_form}) can be factorized into:
\begin{eqnarray}
\mathcal{H}&=&\pi(\eta_1)\times\pi(\eta_2|[\eta_1])...\times\pi(\eta_M|[\eta_1,...,\eta_{M-1}])\notag\\
&=&\pi(\eta_1)\prod_{m=2}^M\pi(\eta_m|[\eta_1,...,\eta_{m-1}])\label{eqn_decoder_factor}
\end{eqnarray}
Therefore it is a natural idea to employ a RNN as the decoder to produce both $\eta_m$ and $\pi(\eta_m|[\eta_1,...,\eta_{m-1}])$, given $[\eta_1,...,\eta_{m-1}]$ where order matters.

\begin{figure}[hbt]
    \begin{center}
        \begin{minipage}{0.85\linewidth}
            \begin{center}
                \includegraphics[width=\columnwidth]{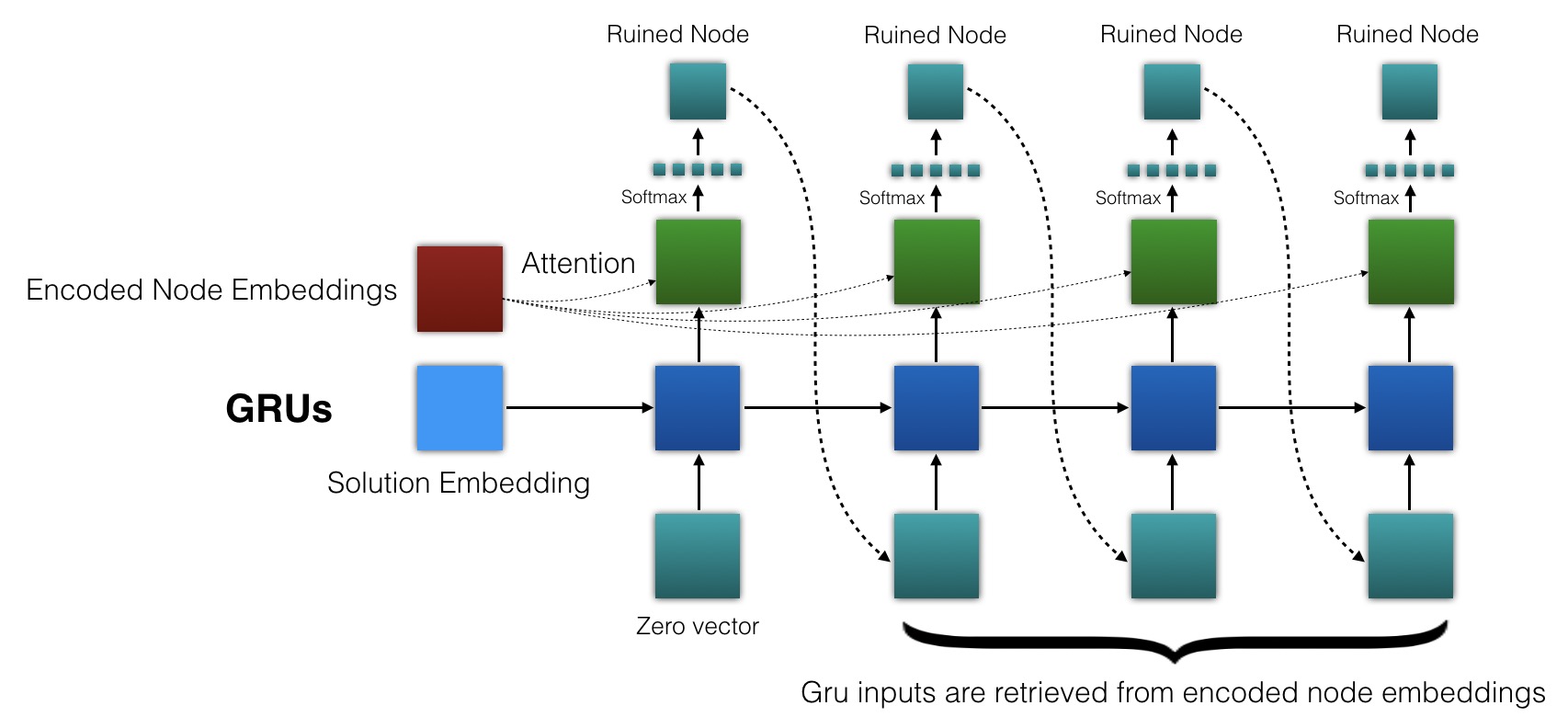}
            \end{center}
        \end{minipage}
    \end{center}
    \caption{The network structure of the decoder.}\label{fig:decoder}
\end{figure}

The decoder in our approach follows the decoder design of Pointer Network, as shown in Fig. \ref{fig:decoder}. The sequential input to each GRU is the node embedding of the candidate node chosen from previous step, i.e. $n_{\text{EGATE},\eta_{m-1}}$ where $n_{\text{EGATE},.}$ is the node embedding defined in (\ref{eqn_egate_embedding}), $\eta_{m-1}$ is the node chosen by previous step. The output of GRU is applied to the attention mechanism with embeddings for all nodes, followed by a softmax layer that produces the probability. The candidate node $\eta_m$ is then selected either in a greedy way or by sampling, together with a conditional probability $\pi(\eta_m|[\eta_1,...,\eta_{m-1}])$ developed by the softmax layer, and is fed to the next GRU unit.

The decoder serves as the destroy operator by producing the nodes to be removed, each of which with a probability, and serves as the repair operator as well, by following the sequence order of produced nodes.

\subsection{Train the Network}
We apply the actor-critic framework of Reinforcement Learning to training the parameters of the encoder and the decoder. Denote the parameter set as $\theta$ for encoder and decoder. At step $t$, the reward is defined as the reduction of VRP cost function, where the VRP cost function is the sum of total traveling distance and the cost of vehicles, as shown below:
\begin{eqnarray}
Cost_{VRP}^{(t)}&=&Distance^{(t)}+C\times K^{(t)}\notag\\
r^{(t)} &=& Cost_{VRP}^{(t)}-Cost_{VRP}^{(t-1)} \label{eqn_reward}
\end{eqnarray}
where $K^{(t)}$ is the number of vehicles being assigned at step $t$, $C$ is the weight of vehicle costs, and $r^{(t)}$ is the corresponding reward.

The value network takes the output of encoder $Enc^{(t)}$ as the \textit{state} at step $t$, and estimates the state value as $\hat{v}(Enc^{(t)}, \phi)$ where $\phi$ is the parameter set of the value network. In our practice the value network is a two-layered feed-forward neural network, where the first layer is a dense layer with ReLU activation and the second layer is a linear one. The training process of actor-critic cycle is shown as: (1) to calculate Advantage as the TD error $\delta_{\text{TD}}$:
\begin{equation}
\delta_{\text{TD}}^{(t)}\leftarrow r^{(t)}+\gamma\hat{v}(Enc^{(t)}, \phi)-\hat{v}(Enc^{(t-1)}, \phi)\label{eqn_td}
\end{equation}
(2) to train the critic network:
\begin{equation}
\phi \leftarrow \phi+\alpha_\phi\delta_{\text{TD}}^{(t)}\nabla_\phi\hat{v}(Enc^{(t)}, \phi)\label{eqn_train_phi}
\end{equation}
(3) to train the actor via clipped surrogate objective Proximal Policy Optimization (PPO) method \cite{schulman2017proximal}:
\begin{equation}
L^{CLIP}(\theta) = \hat{\mathbb{E}}_t[\text{min}(r_t(\theta)\delta_{\text{TD}}^{(t)}, \text{clip}(r_t(\theta), 1-\epsilon, 1+\epsilon)\delta_{\text{TD}}^{(t)})]\label{eqn_train_theta}
\end{equation}
where $Enc^{(t)}$ is the output of encoder at step $t$, $\pi()$ is defined in (\ref{eqn_decoder_factor}), $r^{(t)}$ is the reward defined in (\ref{eqn_reward}), $\alpha_\phi$ is the learning rate for critic, and $\gamma$ is the discount factor, $r_t(\theta)$ is the ratio of new policy over old policy, and $\epsilon=0.2$.

At each step, the updated solution $x^{(t)}$ produced by the decoder is to substitute the previous solution $x^{(t-1)}$ if less vehicles are required, or the total traveling distance satisfies:
\begin{equation}
Distance^{(t)}<Distance^{(t-1)}-T^{(t)}*\log(Rnd)
\end{equation}
where $Rnd$ is a random number uniformly distributed within $[0,1]$, and $T^{(t)}$ is the temperature of Simulated Annealing (SA). This temperature is faded in a way as $T^{(t)}=\alpha_TT^{(t-1)}$.

The entire algorithm is described as below.

\begin{algorithm}
    \caption{VLNS with Learned Heuristics}
    \label{alg:learned_heuristics}
    \begin{algorithmic}
       \STATE {\bfseries Initialization:} $i_{epoch}=0$
       \REPEAT
       \STATE Initialize dataset $\mathcal{G}=\{\mathcal{N},\mathcal{A}\}$, temperature $T$ for SA;
       \STATE Construct an initial feasible solution $x^{(0)}$, set $x^\ast=x^{(0)}$;
       \FOR{$t=1$ {\bfseries to} $N$}
       \STATE Updated solution from encoder-decoder as $x^{(t)}$
       \IF{in learning mode}
       \STATE Calculate reward $r^{(t)}$ from (\ref{eqn_reward})
       \STATE Calculate $\delta_{TD}^{(t)}$ from (\ref{eqn_td})
       \STATE Train actor and critic parameters $\theta, \phi$ as in (\ref{eqn_train_theta}) and (\ref{eqn_train_phi})
       \ENDIF
       \IF{$x^{(t)}$ better than $SA(x^{(t-1)})$}
       \STATE Apply the updated solution $x^{(t)}$
       \STATE Update $x^\ast$ if {$x^{(t)}$} better than $x^\ast$
       \ENDIF
       \ENDFOR
       \UNTIL{$i_{epoch}>N_{epoch}$}
    \end{algorithmic}
\end{algorithm}

\section{Experiments}

\subsection{Setup}
We have performed experiments for two VRP set-ups: the basic CVRP (VRP with constraint on vehicle capacity) and the widely adopted CVRPTW (VRP with constraints on vehicle capacity and service time window).
\subsubsection{\textbf{CVRP}}

In each instance, 100 nodes are randomly generated on a $100*100$ map, where the first node is the depot, and the rest are customer nodes. The demand for each node is an uniformly distributed random variable within $[1,9]$. The vehicles' capacity is 100. Because there is no service time window constraints for CVRP, the primitive node embedding for this case contains only 5 dimensions, by eliminating elements (1), (2) and (8) in section 3.1.

\subsubsection{\textbf{CVRPTW}}

The training setups are the same to CVRP except for the time window. The starting time is uniformly distributed between $T_{start,min}$ = 0, and $T_{start,max}$ = 290. The due time is uniformly distributed between $T_{due,min}$ = 10, and $T_{due,max}$ = 300. The service time period is 10. The depot requires no service time and its time window is $[0,300]$. Vehicles travel a unit distance per unit time.

\subsection{Training Details}

We apply the above settings to generate the training and test sets. We trained the models for 1000 epochs in total, and every training epoch contains 128 randomly generated instances. (Note the training data is generated in parallel to and independently with the training process. Each instance contains $N_{rollout}$ roll-outs, and for each roll-out there are $m$ training samples generated as the $k$-step TD errors, $k=1,...,m$ respectively. It says, there are $128*m*N_{rollout}$ training samples in every epoch.) The test set contains 100 instances which is used to compare the performance of baselines and our model.

The encoder consists of $L_E$ layers of EGATE with hidden size $N_E$, that maps the 8-dimensional primitive node embedding into $N_E$-dimensional extended node embedding, and maps the 2-dimensional primitive edge embedding into $N_{I,edge}$-dimensional extended edge embedding. The output of each EGATE layer is also an vector of $N_E$ dimensional. At the decoder side, the hidden size of cells is $N_D$. The Critic is a fully connected neural network contains one hidden layer with a size of $N_C$. The output of the Critic is a scalar variable representing the value of the current state.

For both tasks, the training batch size $BS= 64, m=10, N_{rollout}=20, L_E=2, N_E= N_D = N_C = 64$, and $N_{I,edge} = 16$. These hyper-parameters are chosen for the purpose to train the models within 100 hours. The algorithms are running on a server with 2 Nvidia 2080Ti GPUs and 40 CPU cores. Only one GPU is used when evaluating the neural network. All neural networks in our evaluation are implemented in PyTorch \cite{paszke2017automatic}. The model is trained with Adam optimizer, the learning rate is $3e-4$, and evaluation batch size is 100 for CVRP task and 256 for CVRPTW task, as explained in next section.

\subsection{Experiments Results}
We compare our algorithm with both handcrafted heuristics designs and neural combinatorial optimization algorithms. The handcrafted heuristics designs include three approaches: \textit{Random} where the heuristics operator is a random function, \textit{Adaptive Large Neighbourhood Search (ALNS)} \cite{Ropke2006AnAL} that chooses the best heuristics at runtime in an adaptive way, and \textit{Slack Induction by String Removals (SISR)} \cite{ChristiaensJan2019SIbS} where nodes are removed based on the combination of current routes and distances between nodes. For neural combinatorial optimization algorithms, we managed to reproduce the \textit{Attention model (AM)} \cite{Kool2018AttentionLT}. We tried two types of settings, either by using greedy search for decoding, or by applying the best result from a batch of sampled results as the softmax results.

We have run 1000 iterations for all models except for SISR that is with 1,000, 200,000, and 1 million iterations, gradually. The running result of SISR with 1 million iterations can be used as a benchmark close enough to the global optimal. The experiments results are shown in table \ref{tbl_cvrp} for CVRP and in table \ref{tbl_cvrptw} for CVRPTW. The model names in the tables are in form of \{model name\}[evaluation batch size]-\{iteration number\}, e.g., AM1280 stands for Attention Model with evaluation batch size 1280, and EGATE100-1K stands for our proposed model with evaluation batch size 100 and 1000 iterations in total. The costs in these tables are the averaged one over 100 instances. If tested in batch, i.e., batch size $>1$, the minimum cost of the batches per instance is chosen as the result for each instance. The results show that, for CVRP, our approach with 1000 iterations is very close to the benchmark with a gap of only 0.58\%, and outperforms all algorithms with 1000 iterations; for CVRPTW, out approach is with the best result with 1000 iterations, and outperforms the benchmark generated by SISR with 1 million iterations.

\begin{table}[h]
\centering
\begin{minipage}[t]{0.32\linewidth}
    \begin{tabular}{ |c|c| } 
    \hline
    Model Name &  Average Cost \\ 
    \hline
    Random-1K &  1188.14 \\ 
    ALNS-1K &  1163.63 \\ 
    \hline
    SISR-1K &  1140.38 \\ 
    SISR-200K &  1074.65 \\ 
    SISR-1M &  $\mathbf{1071.91}$ \\ 
    \hline
    AM1280 & 1144.64 \\ 
    AM-Greedy &  1189.76 \\ 
    \hline
    EGATE-1K &  1148.79 \\ 
    EGATE100-1K &  $\mathbf{1078.16}$ \\ 
    \hline
    \end{tabular}
    \caption{CVRP performance test between different solvers}\label{tbl_cvrp}
\end{minipage}\hfill
\begin{minipage}[t]{0.32\linewidth}
    \begin{tabular}{ |c|c| } 
    \hline
    Model Name &  Average Cost \\ 
    \hline
    Random-1K &  2567.85 \\ 
    ALNS-1K &  2533.50 \\ 
    \hline
    SISR-1K &  2584.69 \\ 
    SISR-200K &  2421.45 \\ 
    SISR-1M &  2419.01 \\ 
    \hline
    EGATE-1K &  2537.23 \\ 
    EGATE256-1K &  $\mathbf{2415.16}$ \\ 
    \hline
    \end{tabular}
    \caption{CVRPTW performance test between different solvers}\label{tbl_cvrptw}
\end{minipage}\hfill
\begin{minipage}[t]{0.32\linewidth}
    \begin{tabular}{ |c|c| } 
    \hline
    Model Name &  Average Cost \\ 
    \hline
    Random-1K &  7622.97 \\ 
    ALNS-1K &  8095.00 \\ 
    SISR-1K &  7900.75 \\ 
    SISR-1M &  $\mathbf{6630.10}$ \\ 
    \hline
    EGATE-1K &  7146.05 \\ 
    EGATE192-1K &  $\mathbf{6924.70}$ \\ 
    \hline
    \end{tabular}
    \caption{CVRPTW (400 nodes) performance test between our proposed approach and other handcrafted heuristic solvers}\label{tbl_400_CVRPTW}
\end{minipage}
\end{table}

\begin{figure}[htb]
    \begin{center}
        \begin{minipage}{0.48\linewidth}
            \centering
            \includegraphics[width=.98\linewidth]{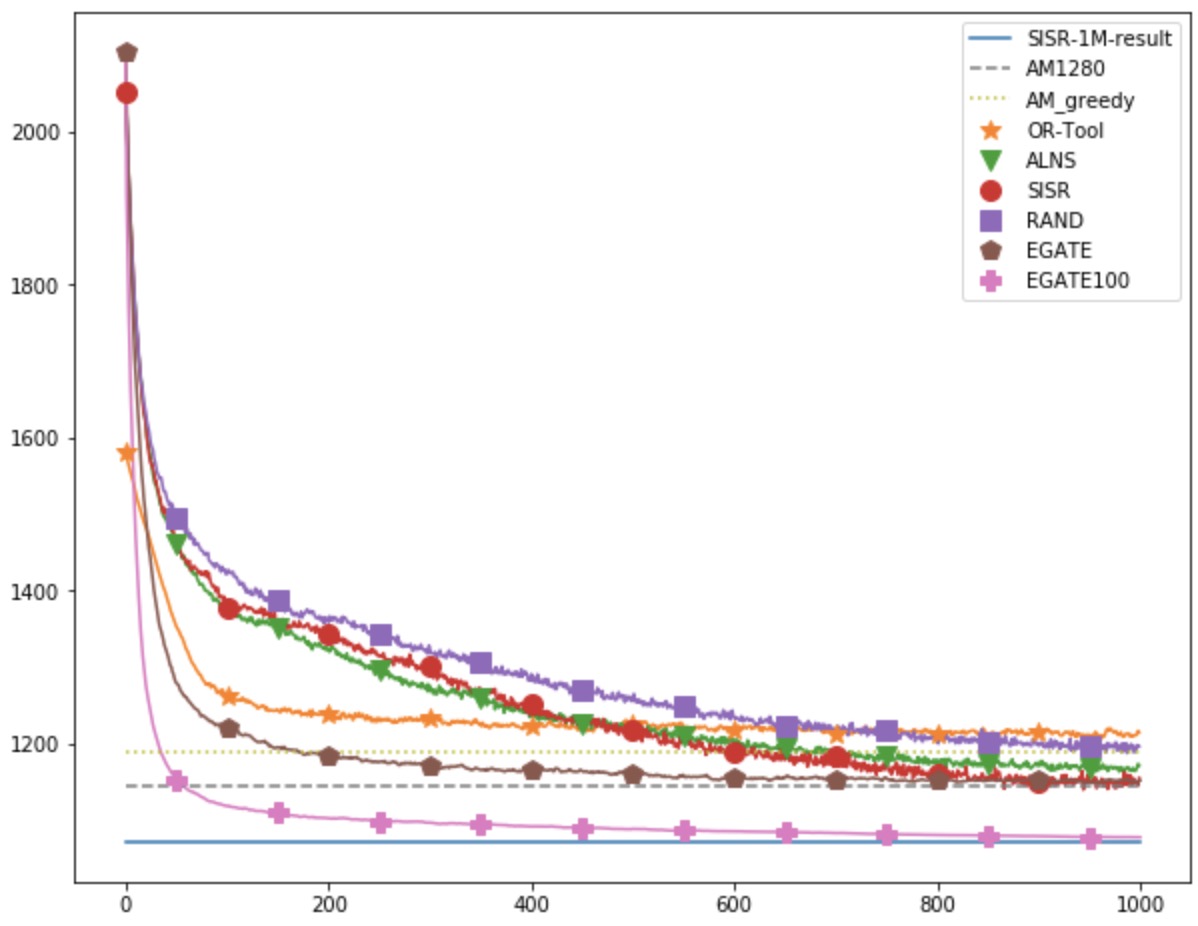}
        \end{minipage}\hfill
        \begin{minipage}{0.48\linewidth}
            \centering
            \includegraphics[width=.98\linewidth]{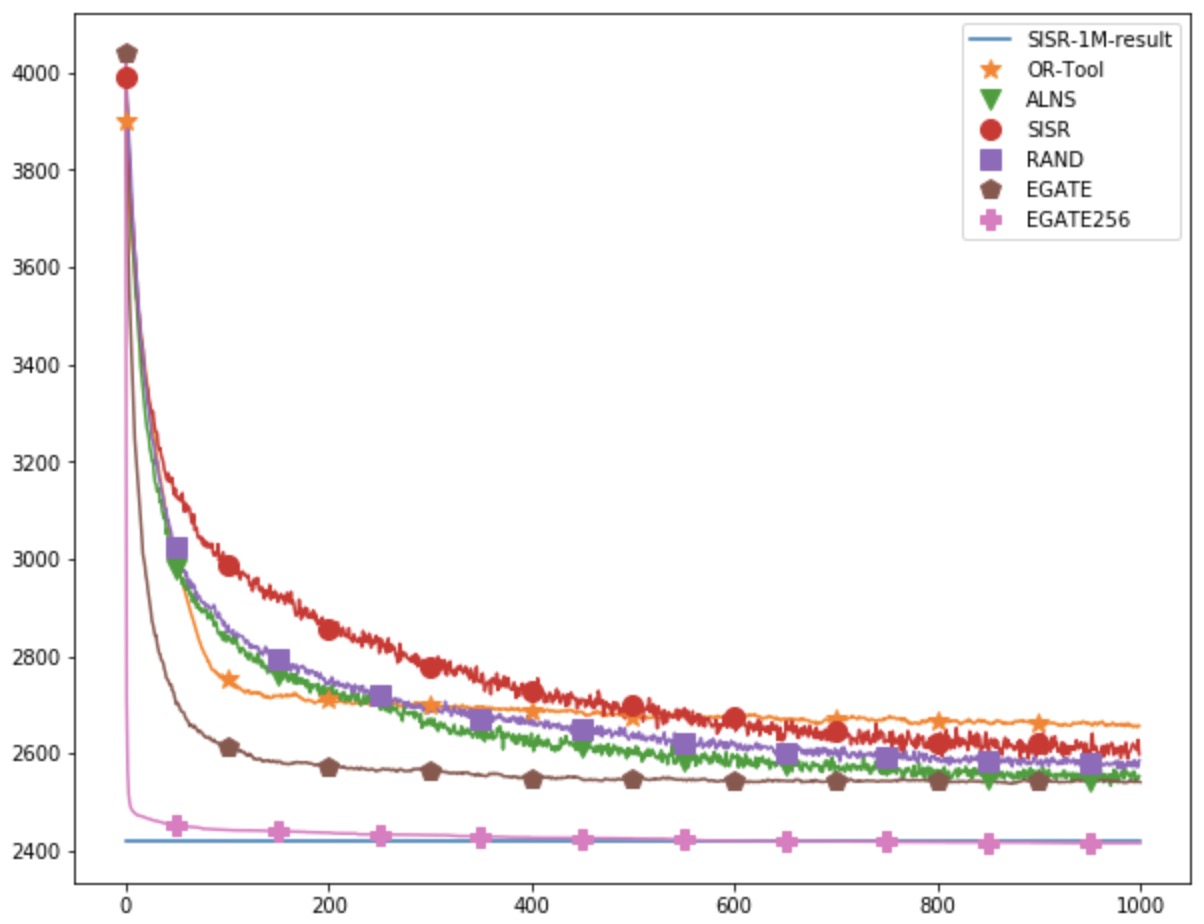}
        \end{minipage}
    \end{center}
    \caption{Left: Comparison for CVRP with different models; Right: Comparison for CVRPTW with different models}\label{fig_cvrp_cvrptw_converge}
\end{figure}

Fig \ref{fig_cvrp_cvrptw_converge} illustrate how the average costs converge along the iterations for the above algorithms. The curves show that our proposed algorithm converges to the benchmark much faster than any other algorithms under test. Meanwhile evaluation with batch is beneficial for both the final results and converging rates. The intuitive interpretation is that batching leads to better exploration of the neighborhood. This statement is pending for further research though.

\subsection{Solving Large-Scale Problems}
VRP, especially CVRPTW, with large-scale data set, for instance over 400 nodes, is a known and considerable challenge in combinatorial optimization. There is no reported experiment results by known neural combinatorial optimization algorithms with over 400 nodes, therefore we compare our approach with handcrafted heuristic algorithms only. Each instance contains 400 randomly generated nodes (including a depot as the 0-th one) in a similar way as described in the setup section. The hyper-parameters keep unchanged except for test batch size as 192 and $L_E=3$, i.e., 3 EGATE layers are stacked to improve the receptive field as the topology scales up. To simplify the computation, in each EGATE layer and for each node $i$, all nodes and related edges other than those who are within top 10\% nearest to the node $i$, as well as the existing nodes in current routes and the depot itself, are masked in the EGATE layers. In this way the whole topology becomes sparse, yet more EGATE layers guarantee the spatial exploration. In this setup 32 CPU cores are used. The experiment results are shown in the table \ref{tbl_400_CVRPTW} and fig (\ref{Fig:vrptw400points}) below. From the table and figure we can see that our proposed approach outperforms ALNS and SISR within the same iterations, and is close to the benchmark with a gap of 4.4\%

\begin{figure}[hbt]
\begin{center}
    \begin{minipage}{0.75\linewidth}
        \begin{center}
            \includegraphics[width=.95\linewidth]{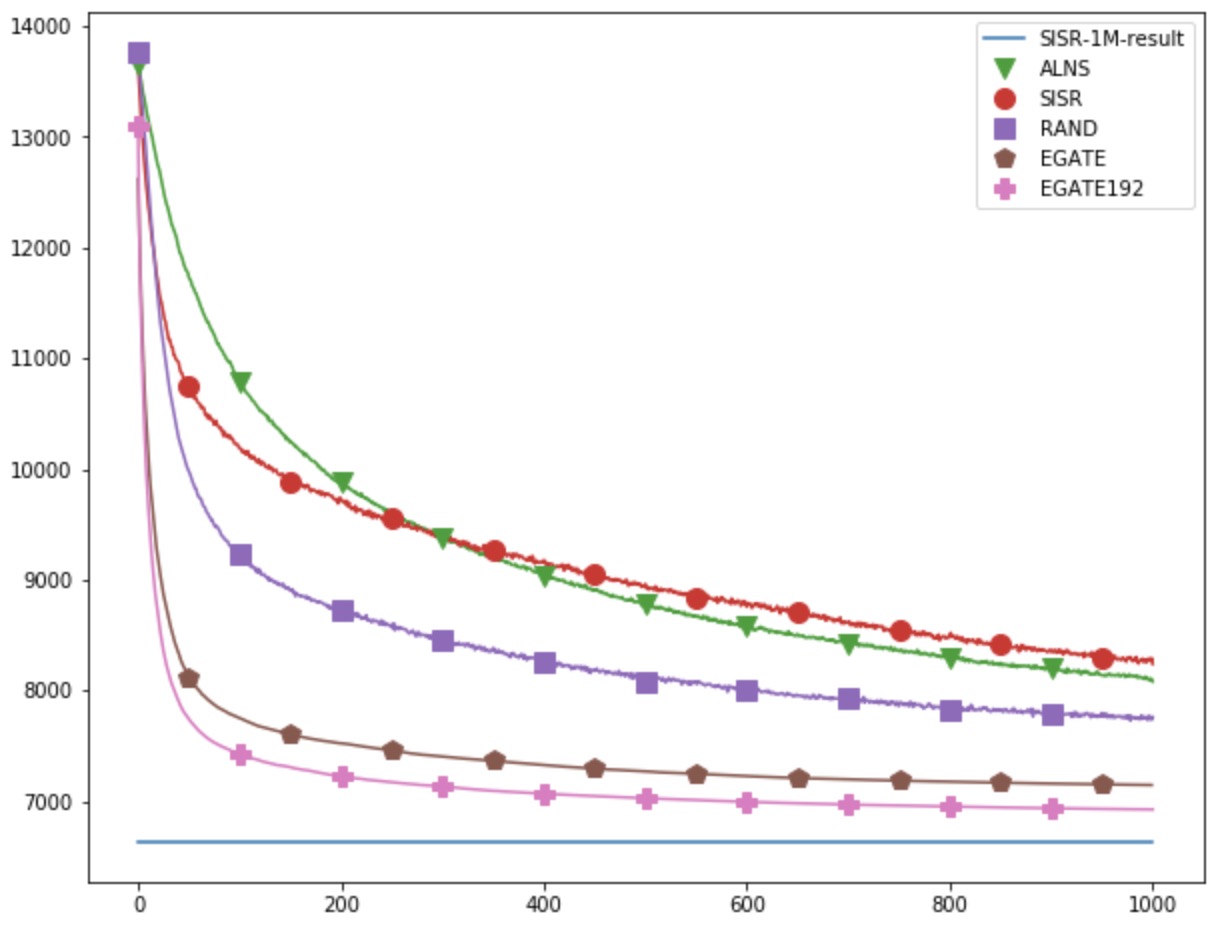}
        \end{center}
    \end{minipage}
\end{center}
\caption{CVRPTW (400 nodes) converging curve}\label{Fig:vrptw400points}
\end{figure}

\section{Conclusion}
In this paper we proposed a novel neural network that learns to design the large-neighborhood search heuristics for vehicle routing problem (VRP). The large-neighborhood search heuristic includes a destroy operator and a repair operator, where the destroy operator removes a set of selected nodes out of current routes, and the repair operator re-inserts these nodes back to other routes to generate an updated solution. We abstract the destroy operator as defining a subset of nodes, and the repair operator as reshaping this subset into an ordered list of the selected nodes, each of which with a probability. In this way the large-neighborhood search heuristic is to produce a stochastic policy $\pi([\eta_1,\eta_2,...,\eta_M])$ where $\eta_i$ is a selected node, and $[.]$ denotes an ordered list. By factorizing this policy into a sequential form as $\pi(\eta_1)\prod_{m=2}^M\pi(\eta_m|[\eta_1,...,\eta_{m-1}])$, it is a natural idea to design and train a neural network model to produce sequential nodes with probabilities, and to train this network with gradient policy architectures such as actor-critic. Our proposed network consists of an encoder and a decoder. The encoder integrates together the node information such as demands and the arc information such as trip costs, based on a modified version of Graph Attention Network (GAT) we presents as EGATE. EGATE considers how to propagate the information carried by the arcs in the graph topology (in form of edge embedding) into the attention weights, together with the information carried by the nodes as proposed by GAT. The EGATE can be stacked into multiple layers to extend the \textit{receptive field} over the graph topology that facilitates the information propagation for nodes and arcs across relatively longer distances. The output of encoder is fed into the decoder to generate sequential output $\pi(\eta_m|[\eta_1,...,\eta_{m-1}])$. Our decoder design is based on GRU and follows the design of Pointer Network. The experiment results on CVRP and CVRPTW with medium-scale data set show that our proposed network outperforms other handcrafted heuristic solvers and other neural combinatorial optimization approaches that we can reproduce, within same iterations. Moreover our proposed network can tackle the large-scale problem tested by date set containing 400 nodes. Our proposed network is the only reported neural combinatorial optimization approach to solve the VRP about this scale to the best of our knowledge, and outperforms other handcrafted heuristic solvers. Therefore our proposed network is capable to learn how to design the large-neighborhood search heuristics, and with a better performance. The need for expertise and handcrafted heuristics design is eliminated due to this fact.

Further research topics include but are not limited to the following suggestions: the network design for other combinatorial optimization problems following this paradigm, especially for the problems where the sequential decisions are to be made; the visibility of EGATE attention weights and the interpretation associated with the graph topology; the applications of EGATE to other graph neural network studies that need to incorporate information from both nodes and arcs, and the masking strategy for different layers of stacked EGATE.

\bibliography{reference}
\bibliographystyle{unsrt}  

\end{document}